\pdfoutput=1
\documentclass[10pt,letterpaper]{article}
\usepackage{amsmath,amsthm,graphicx,booktabs,tabularx,array,placeins}
\IfFileExists{newtxmath.sty}{}{\usepackage{amssymb}}
\usepackage{mmla_technical_report}
\usepackage{orcidlink}
\usepackage{tikz}
\usetikzlibrary{arrows.meta,positioning,fit,calc,backgrounds,shapes.geometric,matrix,decorations.pathreplacing}
\definecolor{Ink}{HTML}{12355B}
\definecolor{Deep}{HTML}{0F172A}
\definecolor{Blue}{HTML}{2F6BFF}
\definecolor{Teal}{HTML}{00A78E}
\definecolor{Orange}{HTML}{F28E2B}
\definecolor{Purple}{HTML}{7B61FF}
\definecolor{Red}{HTML}{D64B4B}
\definecolor{Slate}{HTML}{64748B}
\definecolor{Line}{HTML}{CBD5E1}
\definecolor{Fog}{HTML}{F4F7FA}
\definecolor{Ice}{HTML}{EAF1FF}
\definecolor{Mint}{HTML}{E3F6F1}
\definecolor{Sand}{HTML}{FDF0DD}
\definecolor{Lavender}{HTML}{EFEAFF}
\tikzset{
  flow/.style={-{Latex[length=2.2mm,width=1.6mm]},line width=0.9pt,draw=Ink},
  softflow/.style={-{Latex[length=2.0mm,width=1.5mm]},line width=0.8pt,draw=Slate},
  dashflow/.style={-{Latex[length=2.0mm,width=1.5mm]},line width=0.8pt,dashed,draw=Slate},
  box/.style={rounded corners=2.2mm,draw=Line,line width=0.8pt,fill=white,inner sep=3mm,align=center,font=\sffamily\small,text=Deep},
  smallbox/.style={rounded corners=1.6mm,draw=Line,line width=0.75pt,fill=white,inner sep=2.2mm,align=center,font=\sffamily\footnotesize,text=Deep},
  label/.style={font=\sffamily\scriptsize\bfseries,text=Slate,align=center},
  note/.style={font=\sffamily\scriptsize,text=Slate,align=center},
  title/.style={font=\sffamily\bfseries\large,text=Deep,align=center},
  pill/.style={rounded corners=2mm,inner xsep=2.2mm,inner ysep=1mm,font=\sffamily\scriptsize\bfseries,align=center}
}

\let\originalthebibliography\thebibliography
\renewcommand{\thebibliography}[1]{%
  \originalthebibliography{#1}%
  \setlength{\itemsep}{1pt}%
  \setlength{\parsep}{0pt}%
}

\newtheorem{theorem}{Theorem}[section]

\newtheorem{proposition}[theorem]{Proposition}
\theoremstyle{definition}
\newtheorem{definition}[theorem]{Definition}

\newtheorem{algorithm}[theorem]{Algorithm}
\newtheorem{remark}[theorem]{Remark}

\newcommand{\R}{\mathbb{R}}
\newcommand{\Z}{\mathbb{Z}}
\newcommand{\N}{\mathbb{N}_0}
\newcommand{\card}[1]{\lvert #1\rvert}
\newcommand{\softmax}{\operatorname{softmax}}
\newcommand{\LSE}{\operatorname{LSE}}
\newcommand{\Replace}{\operatorname{Replace}}
\newcommand{\Read}{\operatorname{Read}}
\newcommand{\SoftRead}{\operatorname{SoftRead}}

\newcommand{\Seal}{\operatorname{Seal}}

\hypersetup{
  pdftitle={SoftREU: Differentiable and Numerically Stable Revisable Evidence Updates},
  pdfauthor={Junyi Zou and Avrova Donz},
  bookmarksnumbered=true
}

\title{SoftREU: Differentiable and Numerically Stable\\[-0.06em]
Revisable Evidence Updates}
\author{%
\textbf{Junyi Zou}\textsuperscript{1,*\,\orcidlink{0009-0009-1367-7428}}\quad\textbullet\quad
\textbf{Avrova Donz}\textsuperscript{1,2,*\,\orcidlink{0009-0009-0100-0719}}\\[-0.06em]
{\fontsize{7.3pt}{8.2pt}\selectfont \textsuperscript{1}MMLA-org\quad
\textsuperscript{2}Communication University of China (CUC), No. 1 Dingfuzhuang East Street, Chaoyang District, Beijing 100024, China}\\[-0.10em]
{\small \textsuperscript{*}Equal contribution}\\[-0.10em]
{\scriptsize Email:\enspace
\href{mailto:lior.j.zou@gmail.com}{\nolinkurl{lior.j.zou@gmail.com}}
\enspace$\cdot$\enspace
\href{mailto:avrovadonz@icloud.com}{\nolinkurl{avrovadonz@icloud.com}}}}
\date{November 30, 2026}

\begin{document}
\makemmlatitle{
Revisable Evidence Update (REU) lifts multiplicative Bayesian conditioning from the log semiring into the abelian group $(\Z\times\R)^{\Omega}$ with masked softmax readout, separating each world's zero multiplicity $n_z$ from its positive log mass $b_z$ so named evidence can be withdrawn and blocked support can recover. Embedding this algebra in a learned memory system raises three algorithmic obstacles: hard zero masks block gradient flow and return $\bot$ under total contradiction, direct log-domain subtraction suffers catastrophic cancellation in IEEE-754 floating-point arithmetic, and continuous addressing gradients must respect discrete commit authorization without misleading learning. We introduce \textbf{SoftREU} under a single unified ledger invariant. First, SoftREU replaces hard masking with the tempered readout $p_\beta = \softmax(b - \beta n)$, which uniquely minimizes a conflict-penalized variational free energy, converges exponentially to exact REU as $\beta\to\infty$ on consistent hard-encoded ledgers, concentrates on minimum-conflict worlds under zero total mass, and obeys a tight $\tanh(\epsilon/2)$ total-variation stability bound under log-sum perturbations uniformly over all $\beta > 0$; for smooth factor encodings during training, we prove that $\epsilon \ge e^{-\beta}$ characterizes score monotonicity. Second, a subtraction-free segment tree over handle slots maintains the aggregate in $O(\card{\Omega}\log H)$ additions per slot update, matching a fixed-tree rebuild bitwise without drift from withdrawn factors, supporting atomic block deltas, and folding immutable records into the base aggregate with an explicit end-to-end rounding and total-variation bound. Third, we formulate commit-gated straight-through training alongside the exact score gradient of expected committed loss: we derive a sharp Popoviciu curvature remainder and sufficient sign-agreement condition, and construct a three-world, two-withdrawal witness on which both deterministic and sampled straight-through updates point uphill even with strictly positive likelihoods and exact arithmetic, while a later authorized withdrawal reverses the branch preference. Executable checks verify the analytical bounds, binary64 stability, exact-rational protocol and sign-reversal witness, $O(\card{\Omega}\log H)$ scaling up to $H = 16{,}384$ slots, and a synthetic binary-channel fit through tempered reads on fixed zero-masked ledgers.}

\section{Introduction}\label{sec:intro}

Revisable Evidence Update (REU) organizes exact late evidence revision into three composable stages: an authenticated record transition $\pi_{\mathrm{rec}}$, an algebraic replacement kernel in the abelian group $(\Z\times\R)^{\Omega}$, and a masked softmax query read. By separating zero multiplicities $n_z$ from positive log masses $b_z$, REU avoids the absorbing-zero obstruction of the log semiring $(\R\cup\{-\infty\},+)$ and keeps incremental named edits exact relative to active-ledger reconstruction in real arithmetic.

Embedding REU as a memory and readout carrier inside a learned architecture such as Memory-Mediated Learning Architecture (MMLA)~\cite{mmla-v4} exposes three algorithmic bottlenecks that exact arithmetic does not solve:
\begin{enumerate}
  \item \textbf{Non-differentiable support switching and undefined zero-mass reads.} The hard mask $\mathbf{1}_{\{n_z=0\}}$ has zero or undefined derivatives across support transitions ($n_z: 1 \leftrightarrow 0$) and forces exact REU to return $\bot$ whenever every world carries at least one blocking factor ($U_K = \varnothing$). A learned observation or addressing model receives no gradient across these boundaries.
  \item \textbf{Floating-point catastrophic cancellation.} Direct incremental updates subtract removed log terms from a running sum $b_z$. In IEEE-754 binary64 arithmetic, adding a large negative log factor (such as $-2^{53}$) alongside a moderate term ($-1$) and later withdrawing the large term destroys the smaller contribution irreversibly.
  \item \textbf{Tension between continuous addressing gradients and discrete commit authority.} Mutating a live ledger requires discrete handle attribution, generation validation, visibility consistency, and authenticated grants. Yet even when candidate edits are gated and previewed on isolated copies, differentiating a readout at one committed branch through mixed preview logits can disagree in sign with the expected committed loss across legal branches.
\end{enumerate}

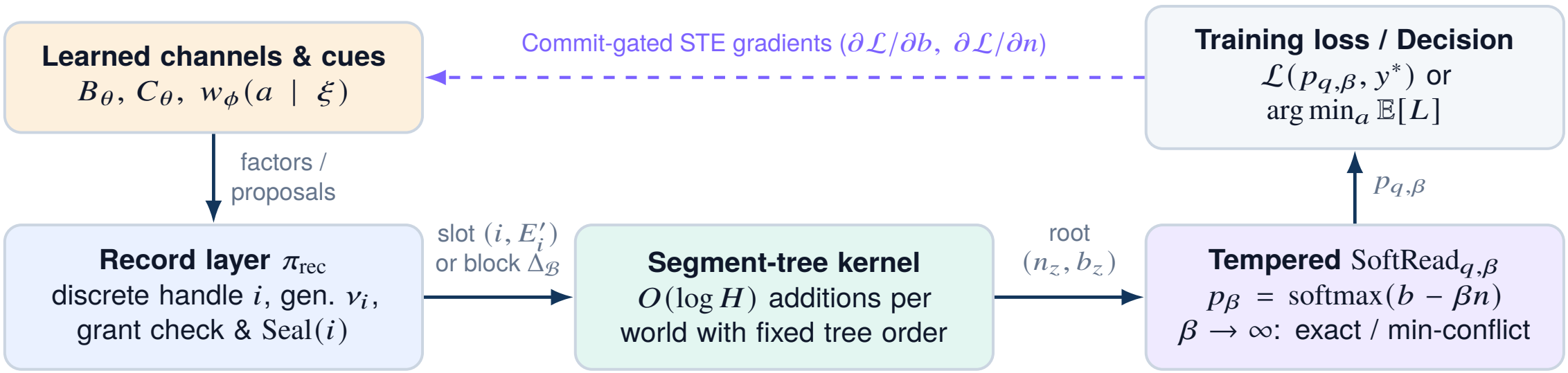
\begin{figure}[tbp]
\centering
\begin{tikzpicture}[x=1cm,y=1cm,
  stage/.style={smallbox,text width=3.55cm,minimum height=1.38cm},
  inputstage/.style={smallbox,fill=Sand,text width=3.55cm,minimum height=1.00cm},
  grad/.style={-{Latex[length=2.1mm,width=1.5mm]},line width=0.85pt,dashed,draw=Purple},
  lbl/.style={font=\sffamily\scriptsize,text=Slate,align=center}]

\node[inputstage] (obs) at (0, 2.10)
  {\textbf{Learned channels \& cues}\\$B_\theta,\, C_\theta,\ w_\phi(a\mid\xi)$};

\node[stage,fill=Ice] (rec) at (0, 0)
  {\textbf{Record layer $\pi_{\mathrm{rec}}$}\\discrete handle $i$, gen.\ $\nu_i$,\\grant check \& $\Seal(i)$};

\node[stage,fill=Mint] (tree) at (5.45, 0)
  {\textbf{Segment-tree kernel}\\$O(\log H)$ additions per\\world with fixed tree order};

\node[stage,fill=Lavender] (read) at (10.9, 0)
  {\textbf{Tempered $\SoftRead_{q,\beta}$}\\$p_\beta = \operatorname{softmax}(b-\beta n)$\\$\beta\to\infty$: exact / min-conflict};

\node[inputstage,fill=Fog] (loss) at (10.9, 2.10)
  {\textbf{Training loss / Decision}\\$\mathcal{L}(p_{q,\beta}, y^*)$ or $\arg\min_a \mathbb{E}[L]$};

\draw[flow] (obs) -- node[lbl,right,xshift=1pt] {factors /\\proposals} (rec);
\draw[flow] (rec) -- node[lbl,above,yshift=2pt] {slot $(i, E'_i)$\\or block $\Delta_{\mathcal B}$} (tree);
\draw[flow] (tree) -- node[lbl,above,yshift=2pt] {root\\$(n_z, b_z)$} (read);
\draw[flow] (read) -- node[lbl,right,xshift=1pt] {$p_{q,\beta}$} (loss);

\draw[grad] (loss.west) -- node[lbl,above,yshift=2pt,text=Purple] {Commit-gated STE gradients ($\partial \mathcal{L}/\partial b,\ \partial \mathcal{L}/\partial n$)} (obs.east);

\end{tikzpicture}
\caption{Architecture of SoftREU. The authenticated record layer $\pi_{\mathrm{rec}}$ validates discrete handles, generations, visibility, and grants, feeding updated effective slot contributions $E'_i$ or atomic block deltas $\Delta_{\mathcal B}$ into a subtraction-free segment tree over $(\N\times\R)^{\Omega}$. The tempered readout $p_\beta = \softmax(b-\beta n)$ supplies smooth gradients during training (dashed path) and recovers exact masked softmax as $\beta\to\infty$ on hard-encoded consistent ledgers.}
\label{fig:softreu-arch}
\end{figure}

This paper introduces \textbf{SoftREU}, upgrading each stage of REU while preserving a unified ledger invariant across edits, visibility changes, and record folding (Figure~\ref{fig:softreu-arch}):

\begin{itemize}
  \item \textbf{Tempered log-group primitive (Section~\ref{sec:softreu}).} Replacing the hard zero mask with $p_\beta(z) = \softmax(b - \beta n)_z$ yields a smooth readout with shift invariance and closed-form Jacobians (Proposition~\ref{prop:gradients}). We prove that $p_\beta$ uniquely minimizes a conflict-penalized variational free energy (Proposition~\ref{prop:variational}), reduces expected conflict monotonically with $\beta$, and satisfies a tight $\tanh(\epsilon/2)$ total-variation bound under $\ell_\infty$ log-sum perturbations uniformly across all $\beta>0$ (Proposition~\ref{prop:perturbation-stability}). As $\beta\to\infty$, SoftREU recovers exact REU at an explicit exponential $L_1$ rate on consistent hard-encoded ledgers and converges to the normalized positive-log posterior over \emph{minimum-conflict worlds} $\arg\min_z n_z$ when exact REU halts with $\bot$ (Theorem~\ref{thm:beta-limit}). For smooth factor encodings during training, we prove that $\epsilon \ge e^{-\beta}$ is necessary and sufficient for score monotonicity (Remark~\ref{rem:soft-zero-mono}).
  \item \textbf{Subtraction-free segment-tree aggregation and safe folding (Section~\ref{sec:segtree}).} Storing the base aggregate and per-slot effective contributions in a balanced binary tree eliminates floating-point subtraction from aggregation. Each single-slot update executes in $O(\card{\Omega}\log H)$ pure additions and matches a fixed-tree rebuild bitwise, leaving zero residual drift from withdrawn transient factors (Theorem~\ref{thm:segtree}). Coordinated multi-handle updates reduce to atomic group deltas $\Delta_{\mathcal B}$, and a $\Seal(i)$ transaction folds immutable visible contributions into the base aggregate with explicit binary64 log-sum and posterior total-variation reassociation bounds.
  \item \textbf{Commit-gated learning and gradient alignment theory (Section~\ref{sec:learning}).} Commit-gated straight-through training (Algorithm~\ref{alg:commit-gated}) preserves the exact ledger invariant and hard-encoded inference limit on every live commit (Proposition~\ref{prop:commit-soundness}). Comparing its score-gradient surrogate against the exact gradient of expected committed loss over admissible branches, we prove a closed-form curvature remainder and sharper sign-agreement condition (Theorem~\ref{thm:ste-curvature}), and construct an exact three-world, two-withdrawal witness on which both deterministic and sampled straight-through updates increase expected committed loss despite strictly positive likelihoods and exact arithmetic---and on which a later authorized withdrawal reverses the preferred branch (Proposition~\ref{prop:ste-reversal}).
\end{itemize}

Section~\ref{sec:experiments} verifies the convergence bound, Jacobians, monotonicity and folding boundaries, binary64 cancellation resistance, a synthetic binary-channel fit on fixed zero-masked ledgers, exact-rational protocol and sign-reversal checks, and single-slot update scaling up to $H = 16{,}384$ slots through reproducible ancillary scripts under \texttt{anc/checks/}. Section~\ref{sec:discussion} situates SoftREU relative to softmax relaxations, straight-through estimators, arithmetic circuits, truth maintenance, and MMLA.

\section{Background: Exact REU and Its Algorithmic Barriers}\label{sec:background}

We formulate exact Revisable Evidence Update (REU) on a nonempty finite world set $\Omega$ of cardinality $N = \card{\Omega}$, a world prior $\rho:\Omega\to\R_{\ge 0}$ with $\sum_{z\in\Omega}\rho(z) = 1$, a finite query family $\mathcal Q$ of maps $q:\Omega\to\mathcal Y_q$, and a finite handle space $\mathcal H$.

\paragraph{Log-group encoding and guarded readout.}
Each scalar factor value $f\ge 0$ is encoded into the additive monoid $\N\times\R$ by
\[
  I_0(f) = \mathbf{1}_{\{f=0\}}, \qquad
  h(f) = \begin{cases} \log f, & f>0,\\ 0, & f=0,\end{cases} \qquad
  \kappa(F)_z = \bigl(I_0(F(z)),\, h(F(z))\bigr),
\]
for any nonnegative factor $F:\Omega\to\R_{\ge 0}$. In an aggregate $K = (n, b) \in (\N\times\R)^{\Omega}$, $n_z$ counts structural zeros at world $z$ and $b_z$ sums the logarithms of all positive factors at $z$. Replacing a summand $E\in(\N\times\R)^\Omega$ by $E'\in(\N\times\R)^\Omega$ acts in the Grothendieck group completion $(\Z\times\R)^{\Omega}$ via
\begin{equation}
  \Replace(K; E, E') = K - E + E'.
  \label{eq:exact-replace}
\end{equation}
In particular, withdrawing a visible contribution sets $E' = (0,0)$ (equivalently $F' \equiv 1$). Let $U_K = \{z\in\Omega : n_z = 0\}$ denote the active (zero-free) support. The exact guarded read evaluates shifted softmax on $U_K$:
\begin{equation}
  p_K(z) = \begin{cases}
    \dfrac{\exp(b_z)}{\sum_{u\in U_K}\exp(b_u)}, & z\in U_K,\\[1ex]
    0, & z\notin U_K,
  \end{cases}
  \qquad \Read_q(K)(y) = \sum_{z:q(z)=y} p_K(z),
  \label{eq:exact-read}
\end{equation}
returning $\bot$ with status $\mathsf{zeroMass}$ when $U_K = \varnothing$.

\paragraph{Authenticated record layer and unified ledger invariant.}
Equal numerical aggregates can conceal distinct handle assignments: swapping two factors $F_1, F_2$ across handles $i_1, i_2$ leaves $K$ unchanged while altering the posterior after a later named withdrawal of $i_1$. REU therefore wraps the replacement kernel inside an authenticated record layer $\pi_{\mathrm{rec}}$ with control state $C = (R, D, \nu, J, P)$: $R \subseteq \mathcal H$ is the set of live editable handles, $D = (D_i)_{i\in R}$ stores their local payloads, $\nu$ tracks handle generations and retirement or seal status, $J$ records consumed delivery IDs, and $P$ determines the visible subset $V \subseteq \mathcal H$. Each live editable handle $i \in R$ maps deterministically from $D_i$ under the observation model to a nonnegative factor $F_i:\Omega\to\R_{\ge 0}$.

A full evidence state is a tuple $S = (C, K_{\mathrm{base}}, K)$, where $K_{\mathrm{base}} \in (\N\times\R)^{\Omega}$ is an immutable base aggregate initialized to $\kappa(\rho)$ (and augmented only by folding immutable records via $\Seal$ in Section~\ref{sec:segtree}). For every handle $i \in \mathcal H$, define its \emph{effective slot contribution}
\begin{equation}
  E_i(S) = \begin{cases}
    \kappa(F_i), & i \in R \cap V,\\
    (0, 0), & i \notin R \cap V.
  \end{cases}
  \label{eq:effective-slot}
\end{equation}
Thus absent, retired, sealed, and hidden ($i \in R \setminus V$) handles all contribute $(0,0)$. State $S$ is \emph{admissible} if its aggregate satisfies the unified ledger invariant
\begin{equation}
  K = K_{\mathrm{base}} + \sum_{i\in\mathcal H} E_i(S) \;\in\; (\N\times\R)^{\Omega}.
  \label{eq:ledger-invariant}
\end{equation}
When no records have been folded into $K_{\mathrm{base}}$, equation~\eqref{eq:ledger-invariant} reduces to $K = \kappa(\rho) + \sum_{i\in R\cap V}\kappa(F_i)$, matching full active visible ledger reconstruction.

Given an incoming event $e$ specifying a delivery ID $j$, target handle $i$, expected generation, operation (insertion, payload edit, withdrawal, or visibility change), and authorization token, $\pi_{\mathrm{rec}}(S, e)$ atomically checks that $j\notin J$, that handle $i$'s status and generation match $\nu$ (rejecting edits to retired or sealed handles), and that the grant is valid. On rejection, $S$ is unchanged. On acceptance, $\pi_{\mathrm{rec}}$ transitions $C\to C'$ (recording $j$ in $J$, advancing $\nu$, updating $D$, $R$, or $P$, and holding $K'_{\mathrm{base}} = K_{\mathrm{base}}$) and updates the aggregate by replacing the handle's pre-edit effective contribution with its post-edit effective contribution:
\[
  K' = \Replace\bigl(K;\, E_i(S),\, E_i(S')\bigr) = K - E_i(S) + E_i(S').
\]
Because $E_i(S)$ is a summand of $K$ in~\eqref{eq:ledger-invariant} and every other handle $h\neq i$ retains $E_h(S') = E_h(S)$, this transition preserves~\eqref{eq:ledger-invariant} identically in exact arithmetic. In particular, editing a hidden handle $i \in R\setminus V$ has $E_i(S) = E_i(S') = (0,0)$ and leaves $K$ and every query read unchanged until a subsequent publication moves $i$ into $V$, at which point $E_i$ transitions from $(0,0)$ to $\kappa(F_i)$.

\paragraph{Why an algorithmic upgrade is needed.}
Exact REU settles the algebraic semantics of revisable evidence, yet three practical obstacles prevent its direct use in a trainable, finite-precision memory layer:
\begin{enumerate}
  \item \emph{Discontinuity and failure at $U_K=\varnothing$:} The hard mask $\mathbf{1}_{\{n_z=0\}}$ blocks gradient propagation across support changes, and a transient contradiction ($n_z\ge 1$ for all $z$) halts downstream policies with $\bot$ instead of ranking the least-conflicted worlds.
  \item \emph{Floating-point subtraction drift:} Equation~\eqref{eq:exact-replace} subtracts removed log contributions from $b_z$. In finite precision, withdrawing a large-magnitude log factor after accumulating smaller terms permanently corrupts $b_z$.
  \item \emph{Payload growth and discrete addressing:} Retaining all editable record payloads $D_i$ grows with history unless immutable visible records can be folded safely into $K_{\mathrm{base}}$, and discrete handle selection prevents direct backpropagation into learned observation or addressing parameters.
\end{enumerate}
The next three sections resolve these bottlenecks while preserving the invariant~\eqref{eq:ledger-invariant}.

\section{The SoftREU Operator: Tempered Log-Group Relaxation}\label{sec:softreu}

To make revisable evidence updates smooth across support changes and well-defined under transient contradictions without leaving the log-group coordinates $(n,b)$, we replace the hard mask $\mathbf{1}_{\{n_z=0\}}$ with an exponential penalty on zero multiplicity. Throughout, for any $v\in\R^\Omega$, we write $\LSE(v) = \log\sum_{u\in\Omega}\exp(v_u)$.

\begin{definition}[SoftREU readout]\label{def:softreu}
Let $K = (n, b) \in (\R_{\ge 0}\times\R)^{\Omega}$ be a log-group aggregate and let $\beta > 0$ be an inverse conflict temperature. Define the tempered score $s_\beta(z) = b_z - \beta n_z$. The \textbf{SoftREU world distribution} and \textbf{query readout} for $q:\Omega\to\mathcal Y_q$ are
\begin{equation}
  \boxed{\displaystyle
  p_\beta(z) = \softmax(b - \beta n)_z
             = \frac{\exp(b_z - \beta n_z)}{\sum_{u\in\Omega}\exp(b_u - \beta n_u)},
  \qquad
  \SoftRead_{q,\beta}(K)(y) = \sum_{z:q(z)=y} p_\beta(z).}
  \label{eq:softreu-read}
\end{equation}
\end{definition}

Unlike equation~\eqref{eq:exact-read}, $p_\beta(z)$ is strictly positive and well-defined for every finite $(n,b)\in(\R_{\ge 0}\times\R)^{\Omega}$, including states where every world has $n_z > 0$.

\subsection{Exact Limit and Minimum-Conflict Degradation}

SoftREU recovers exact REU exponentially fast whenever a consistent world exists and concentrates on minimum-conflict worlds otherwise.

\begin{theorem}[Exact limit, exponential rate, and minimum-conflict fallback]\label{thm:beta-limit}
Fix $K = (n, b)\in (\N\times\R)^{\Omega}$. Let $n_{\min} = \min_{z\in\Omega} n_z$ be the minimum conflict count and $U_{\min} = \{z\in\Omega : n_z = n_{\min}\}$ be the set of minimum-conflict worlds. Define the limiting distribution
\[
  p_\infty(z) = \begin{cases}
    \dfrac{\exp(b_z)}{\sum_{u\in U_{\min}}\exp(b_u)}, & z\in U_{\min},\\[1.2ex]
    0, & z\notin U_{\min}.
  \end{cases}
\]
Then $p_\beta \to p_\infty$ as $\beta\to+\infty$. Moreover, whenever $U_{\min}\neq\Omega$, letting $\Delta n_{\min} = \min_{z\notin U_{\min}}(n_z - n_{\min}) \ge 1$, the $L_1$ error obeys the non-asymptotic bound
\begin{equation}
  \|p_\beta - p_\infty\|_1
  \;\le\; 2\,\frac{\sum_{z\notin U_{\min}}\exp(b_z)}{\sum_{u\in U_{\min}}\exp(b_u)}\,e^{-\beta \Delta n_{\min}}.
  \label{eq:l1-bound}
\end{equation}
In particular:
\begin{enumerate}
  \item \textbf{Exact REU recovery:} If $U_K = \{z:n_z=0\}\neq\varnothing$, then $n_{\min}=0$, $U_{\min}=U_K$, and $\SoftRead_{q,\beta}(K)$ converges exponentially to $\Read_q(K)$ as $\beta\to\infty$.
  \item \textbf{Minimum-conflict degradation:} If $U_K=\varnothing$ (where exact REU returns $\bot$), $p_\beta$ converges to the normalized positive-log posterior restricted to the least-conflicted worlds $U_{\min}$.
\end{enumerate}
\end{theorem}

\begin{proof}
Subtracting the constant $-\beta n_{\min}$ from every score $b_z - \beta n_z$ leaves $\softmax(b-\beta n)$ invariant, so
\[
  p_\beta(z) = \frac{\exp\bigl(b_z - \beta(n_z - n_{\min})\bigr)}{Z_{\min} + R_\beta},
\]
where $Z_{\min} = \sum_{u\in U_{\min}}\exp(b_u) > 0$ and $R_\beta = \sum_{z\notin U_{\min}}\exp\bigl(b_z - \beta(n_z - n_{\min})\bigr) \ge 0$. For $z\in U_{\min}$, we have $p_\infty(z) = \exp(b_z)/Z_{\min}$ and
\[
  p_\infty(z) - p_\beta(z) = \exp(b_z)\left(\frac{1}{Z_{\min}} - \frac{1}{Z_{\min}+R_\beta}\right)
  = \frac{\exp(b_z)}{Z_{\min}}\cdot\frac{R_\beta}{Z_{\min}+R_\beta} \ge 0.
\]
For $z\notin U_{\min}$, $p_\infty(z)=0$, so $\card{p_\beta(z)-p_\infty(z)} = p_\beta(z)$. Summing over all $z\in\Omega$ yields the exact identity
\[
  \|p_\beta - p_\infty\|_1
  = \sum_{z\in U_{\min}}(p_\infty(z)-p_\beta(z)) + \sum_{z\notin U_{\min}} p_\beta(z)
  = \frac{2 R_\beta}{Z_{\min}+R_\beta} \le \frac{2 R_\beta}{Z_{\min}}.
\]
If $U_{\min} = \Omega$, then $R_\beta = 0$ and $p_\beta = p_\infty$ identically for every $\beta > 0$. Otherwise, $n_z - n_{\min} \ge \Delta n_{\min} \ge 1$ for every $z\notin U_{\min}$, so $R_\beta \le e^{-\beta\Delta n_{\min}}\sum_{z\notin U_{\min}}\exp(b_z)$, establishing~\eqref{eq:l1-bound} and giving $R_\beta\to 0$ as $\beta\to\infty$. Statements~1 and~2 follow by evaluating $n_{\min}=0$ and $n_{\min}\ge 1$.
\end{proof}

Writing $\Read_{q,\infty}(K)(y) = \sum_{z:q(z)=y} p_\infty(z)$ for the limiting query distribution, the triangle inequality over query fibers $q^{-1}(y)$ gives $\|\SoftRead_{q,\beta}(K) - \Read_{q,\infty}(K)\|_1 \le \|p_\beta - p_\infty\|_1$. For any tolerance $\varepsilon \in (0, 2)$, choosing $\beta \ge \log\!\bigl(2\card{\Omega}/\varepsilon\bigr) + (\max_z b_z - \min_z b_z) > 0$ guarantees $\varepsilon$-accuracy on every query.

\subsection{Variational Characterization: SoftREU as Conflict-Penalized Softmax}

Standard softmax maximizes Shannon entropy subject to linear score expectations; SoftREU adds a linear penalty on expected structural conflict.

\begin{proposition}[Variational free-energy characterization]\label{prop:variational}
For any $K=(n,b)\in(\R_{\ge 0}\times\R)^{\Omega}$ and $\beta>0$, let $H(p) = -\sum_{z\in\Omega} p(z)\log p(z)$ (with $0\log 0 = 0$) denote Shannon entropy on the simplex $\Delta(\Omega)$. Then $p_\beta = \softmax(b-\beta n)$ is the unique minimizer of the conflict-penalized variational free energy
\begin{equation}
  \mathcal F_\beta(p;\, n, b)
  \;=\; \beta\,\mathbb{E}_{p}[n_Z] - \mathbb{E}_{p}[b_Z] - H(p),
  \label{eq:free-energy}
\end{equation}
with optimal value $\min_{p\in\Delta(\Omega)} \mathcal F_\beta(p;n,b) = -\LSE(b-\beta n)$.
\end{proposition}

\begin{proof}
For any $p\in\Delta(\Omega)$, summing over the support $\{z : p(z) > 0\}$,
\[
  \mathcal F_\beta(p;n,b)
  = \sum_{z:p(z)>0} p(z)\log\frac{p(z)}{\exp(b_z-\beta n_z)}
  = D_{\mathrm{KL}}(p \,\|\, p_\beta) - \LSE(b-\beta n).
\]
Because $p_\beta(z) > 0$ for all $z\in\Omega$, $D_{\mathrm{KL}}(p \,\|\, p_\beta)$ is finite on $\Delta(\Omega)$, and Gibbs' inequality gives $D_{\mathrm{KL}}(p \,\|\, p_\beta) \ge 0$ with equality if and only if $p = p_\beta$.
\end{proof}

Thus Softmax minimizes $-\mathbb{E}_p[b_Z] - H(p)$ over positive log scores, whereas SoftREU introduces the multiplier $\beta$ to trade expected structural conflict $\mathbb{E}_p[n_Z]$ directly against negative log-evidence and negative entropy.

\subsection{Smooth Gradients, Perturbation Stability, and Soft-Zero Encoding}

\begin{proposition}[Shift invariance and closed-form Jacobians]\label{prop:gradients}
Viewing $p_\beta = \softmax(b-\beta n)$ as a $C^\infty$ map on $(\R\times\R)^\Omega\times\R$ (and hence on $(\R_{\ge 0}\times\R)^\Omega\times(0,\infty)$), for any constants $c_1, c_2\in\R$ we have $\softmax\bigl((b+c_1\mathbf{1}) - \beta(n+c_2\mathbf{1})\bigr) = p_\beta$. Writing $p_{q,\beta}(y) = \SoftRead_{q,\beta}(K)(y)$, the partial derivatives with respect to $b_w$, $n_w$, and $\beta$ are
\begin{align}
  \frac{\partial p_{q,\beta}(y)}{\partial b_w}
    &= p_\beta(w)\bigl(\mathbf{1}_{\{q(w)=y\}} - p_{q,\beta}(y)\bigr),
  \label{eq:grad-b}\\
  \frac{\partial p_{q,\beta}(y)}{\partial n_w}
    &= -\beta\, p_\beta(w)\bigl(\mathbf{1}_{\{q(w)=y\}} - p_{q,\beta}(y)\bigr),
  \label{eq:grad-n}\\
  \frac{\partial p_{q,\beta}(y)}{\partial \beta}
    &= -\operatorname{Cov}_{p_\beta}\!\bigl(\mathbf{1}_{\{q(Z)=y\}},\, n_Z\bigr).
  \label{eq:grad-beta}
\end{align}
\end{proposition}

\begin{proof}
Shift invariance follows from cancelling $\exp(c_1-\beta c_2)$ in numerator and denominator. Since $s_\beta(w) = b_w - \beta n_w$, summing the softmax Jacobian $\partial p_\beta(z)/\partial s_\beta(w) = p_\beta(z)(\mathbf{1}_{\{z=w\}} - p_\beta(w))$ over $z\in q^{-1}(y)$ gives $\partial p_{q,\beta}(y)/\partial s_\beta(w) = p_\beta(w)(\mathbf{1}_{\{q(w)=y\}} - p_{q,\beta}(y))$. Applying the chain rule with $\partial s_\beta(w)/\partial b_w = 1$, $\partial s_\beta(w)/\partial n_w = -\beta$, and $\partial s_\beta(w)/\partial\beta = -n_w$ yields~\eqref{eq:grad-b}--\eqref{eq:grad-beta}.
\end{proof}

\begin{proposition}[Monotone conflict annealing and uniform perturbation stability]\label{prop:perturbation-stability}
Let $K = (n, b) \in (\R_{\ge 0}\times\R)^\Omega$, let $\beta > 0$, and write $\|p - \tilde p\|_{\mathrm{TV}} = \frac{1}{2}\|p - \tilde p\|_1 = \max_{A\subseteq\Omega}\card{p(A) - \tilde p(A)}$ for total-variation distance.
\begin{enumerate}
  \item \textbf{Monotone conflict annealing:} The expected conflict count is nonincreasing in $\beta$,
  \[
    \frac{\partial}{\partial\beta}\mathbb{E}_{p_\beta}[n_Z]
    = -\operatorname{Var}_{p_\beta}(n_Z) \le 0,
  \]
  with strict inequality whenever $n_z$ is not constant on $\Omega$.
  \item \textbf{Uniform log-sum stability:} If approximate log sums $\widehat b \in \R^\Omega$ satisfy $\card{\widehat b_z - b_z} \le \epsilon$ ($\epsilon \ge 0$) for every $z\in\Omega$ while the conflict counts $n$ are unchanged, then $\widehat p_\beta = \softmax(\widehat b - \beta n)$ and $\widehat p_{q,\beta} = \SoftRead_{q,\beta}(n, \widehat b)$ obey
  \begin{equation}
    \|\widehat p_\beta - p_\beta\|_{\mathrm{TV}} \le \tanh\!\left(\frac{\epsilon}{2}\right)
    = \frac{e^\epsilon - 1}{e^\epsilon + 1} \le \frac{\epsilon}{2},
    \qquad
    \|\widehat p_{q,\beta} - p_{q,\beta}\|_{\mathrm{TV}} \le \tanh\!\left(\frac{\epsilon}{2}\right),
    \label{eq:tv-stability}
  \end{equation}
  uniformly over all $\beta > 0$ and all $n \in \R_{\ge 0}^\Omega$. Equality is attained on two worlds by $b=(0,\epsilon)$, $\widehat b=(\epsilon,0)$, and $n=(0,0)$.
\end{enumerate}
\end{proposition}

\begin{proof}
For part~1, applying~\eqref{eq:grad-beta} to the identity query gives $\partial p_\beta(z)/\partial\beta = p_\beta(z)(\mathbb{E}_{p_\beta}[n_Z] - n_z)$, so multiplying by $n_z$ and summing over $z\in\Omega$ yields $\mathbb{E}_{p_\beta}[n_Z]^2 - \mathbb{E}_{p_\beta}[n_Z^2] = -\operatorname{Var}_{p_\beta}(n_Z) \le 0$; because $p_\beta$ has full support on $\Omega$, the variance vanishes if and only if $n_z$ is constant.

For part~2, let $W(z) = \exp(b_z - \beta n_z)$ and $\widehat W(z) = \exp(\widehat b_z - \beta n_z)$, so $e^{-\epsilon}W(z) \le \widehat W(z) \le e^{\epsilon}W(z)$ for all $z\in\Omega$. For any event $A\subseteq\Omega$ with $\alpha = p_\beta(A)\in[0,1]$ and $R = e^{2\epsilon}\ge 1$,
\[
  \widehat p_\beta(A)
  = \frac{\widehat W(A)}{\widehat W(A) + \widehat W(A^c)}
  \le \frac{e^\epsilon W(A)}{e^\epsilon W(A) + e^{-\epsilon} W(A^c)}
  = \frac{R\alpha}{1 + (R - 1)\alpha},
\]
which holds trivially for $A\in\{\varnothing,\Omega\}$ as well. The algebraic identity
\[
  \frac{\sqrt R - 1}{\sqrt R + 1} - \left(\frac{R\alpha}{1 + (R - 1)\alpha} - \alpha\right)
  = \frac{(\sqrt R - 1)\bigl(1 - (\sqrt R + 1)\alpha\bigr)^2}{(\sqrt R + 1)\bigl(1 + (R - 1)\alpha\bigr)} \ge 0
\]
gives $\widehat p_\beta(A) - p_\beta(A) \le (\sqrt R - 1)/(\sqrt R + 1) = \tanh(\epsilon/2)$. Applying the same bound to $A^c$ yields $p_\beta(A) - \widehat p_\beta(A) = \widehat p_\beta(A^c) - p_\beta(A^c) \le \tanh(\epsilon/2)$. Maximizing over $A\subseteq\Omega$ proves the world bound, restricting to pullbacks $A = q^{-1}(E)$ proves the query bound, and $\tanh(u)\le u$ holds for all $u\ge 0$. Direct substitution at $b=(0,\epsilon)$, $\widehat b=(\epsilon,0)$, $n=(0,0)$ yields $p_\beta(0) = (1+e^\epsilon)^{-1}$ and $\widehat p_\beta(0) = e^\epsilon(1+e^\epsilon)^{-1}$, attaining $\tanh(\epsilon/2)$.
\end{proof}

\paragraph{Soft-zero encoding for differentiable factor generators.}
When a learned observation model produces continuous factor probabilities $f_\theta(z)\in[0,1]$ that may approach zero during training, the hard indicator $I_0(f) = \mathbf{1}_{\{f=0\}}$ has zero derivative for $f>0$ and is discontinuous at $0$. For training-time gradient flow into $f_\theta$, we define the smooth encoding with scale $\tau > 0$ and floor $\epsilon > 0$:
\begin{equation}
  \kappa_{\tau,\epsilon}(f)
  = \bigl(\tilde{I}_{0,\tau}(f),\; \tilde{h}_{\tau,\epsilon}(f)\bigr),
  \qquad
  \tilde{I}_{0,\tau}(f) = e^{-f/\tau},
  \qquad
  \tilde{h}_{\tau,\epsilon}(f) = \bigl(1 - e^{-f/\tau}\bigr)\log(f + \epsilon).
  \label{eq:soft-zero}
\end{equation}
At $f=0$, $\tilde{I}_{0,\tau}(0)=1$ and $\tilde{h}_{\tau,\epsilon}(0)=0$, matching $\kappa(0)$ identically; for any fixed $f>0$, $\lim_{\tau\to 0^+,\epsilon\to 0^+}\kappa_{\tau,\epsilon}(f) = (0, \log f) = \kappa(f)$ pointwise, while both coordinates are $C^\infty$ on $f \ge 0$.

\begin{remark}[Monotonicity guard for soft-zero encoding]\label{rem:soft-zero-mono}
Differentiating the single-factor tempered score $s_\beta(f) = \tilde{h}_{\tau,\epsilon}(f) - \beta\tilde{I}_{0,\tau}(f) = (1 - e^{-f/\tau})\log(f+\epsilon) - \beta e^{-f/\tau}$ with respect to $f \ge 0$ yields
\[
  s_\beta'(f)
  = \frac{e^{-f/\tau}}{\tau}\bigl(\beta + \log(f+\epsilon)\bigr)
  + \frac{1 - e^{-f/\tau}}{f+\epsilon}.
\]
For $f>0$ the second term is strictly positive. Hence $\beta+\log\epsilon\ge 0$ guarantees $s_\beta'(f)>0$ for all $f>0$, whereas $\beta+\log\epsilon < 0$ makes $s_\beta'(0)=(\beta+\log\epsilon)/\tau < 0$ and causes the score to decrease immediately to the right of zero. The condition $\epsilon \ge e^{-\beta}$ therefore characterizes score monotonicity on $[0,1]$ (and on all $f \ge 0$); examples include $\epsilon_t = e^{-\beta_t}$ or the capped rule $\epsilon_t = \min(1/2, c e^{-\beta_t})$ with $c \ge 1$ whenever $\beta_t \ge \log 2$. Theorem~\ref{thm:beta-limit} applies to hard-encoded integer counts. For any fixed $f>0$, the relaxed score converges to $\log f$ along a joint schedule if $\epsilon_t\to 0^+$, $\tau_t\to 0^+$, and $\beta_t e^{-f/\tau_t}\to 0$, while matching $\kappa(0)=(1,0)$ at $f=0$.
\end{remark}
\FloatBarrier

\section{Subtraction-Free Aggregation, Block Deltas, and Safe Folding}\label{sec:segtree}

Direct incremental maintenance of $b'_z = b_z - b^{\mathrm{old}}_z + b^{\mathrm{new}}_z$ is exact in real arithmetic, but fails in IEEE-754 floating-point arithmetic. On two equally likely worlds, adding $\log F_1 = (-2^{53}, 0)$ and $\log F_2 = (-1, 0)$ in binary64 with round-to-nearest ties-to-even rounds $-2^{53}-1$ to $-2^{53}$; withdrawing $F_1$ by subtraction leaves a computed subtotal of $0$ instead of $-1$, shifting the first-world posterior from $1/(1+e)\approx 0.2689$ to $1/2$. Removing floating-point subtraction from the aggregation path eliminates this drift.

\subsection{Subtraction-Free Segment-Tree Aggregation}

Fix a handle capacity $H = \card{\mathcal H} \ge 1$ indexing handles $1,\ldots,H$, and let $M = 2^{\lceil\log_2(H+1)\rceil}$ be the number of padded leaves in a complete balanced binary tree. Leaf $M$ (slot $0$) holds the base aggregate $K_{\mathrm{base}}$ (initially $\kappa(\rho)$), leaves $M+1,\ldots,M+H$ hold the effective slot contributions $E_1(S),\ldots,E_H(S)$ of equation~\eqref{eq:effective-slot}, and any padding leaves $M+H+1,\ldots,2M-1$ hold $(0,0)$. Each node $v\in\{1,\ldots,2M-1\}$ stores a per-world pair $(n^{(v)}, b^{(v)})\in(\N\times\R)^{\Omega}$.

\begin{algorithm}[Subtraction-Free Segment-Tree SoftREU]\label{alg:segtree}
Maintain leaf $M$ at $K_{\mathrm{base}}$, leaves $v=M+i$ at $E_i(S)$ for $i\in\{1,\ldots,H\}$, and padding leaves $M+H+1,\ldots,2M-1$ at $(0,0)$.
\begin{enumerate}
  \item \textbf{Update slot $i\in\{0,1,\ldots,H\}$ with new leaf value $E'_i$:} Set leaf $v \leftarrow M + i$ to $(n^{(v)}, b^{(v)}) \leftarrow E'_i$ (where $E'_0 = K'_{\mathrm{base}}$ for the base slot, and $E'_i = E_i(S')$ for handle slot $i\ge 1$). While $v > 1$, set $v \leftarrow \lfloor v/2\rfloor$ and recompute by pure addition:
  \[
    n^{(v)} \leftarrow n^{(2v)} + n^{(2v+1)}, \qquad
    b^{(v)} \leftarrow b^{(2v)} + b^{(2v+1)}.
  \]
  \item \textbf{Read query $q$ at inverse temperature $\beta$:} Read root $(n^{(1)}, b^{(1)})$, compute shifted scores $\tilde{s}_z = (b^{(1)}_z - \beta n^{(1)}_z) - \max_u(b^{(1)}_u - \beta n^{(1)}_u)$~\cite{Blanchard_lse}, and return $\SoftRead_{q,\beta}$ via normalized $\exp(\tilde{s}_z)$.
\end{enumerate}
\end{algorithm}

\begin{theorem}[Exactness, update cost, and fixed-tree history independence]\label{thm:segtree}
Starting from an admissible state $S$ initialized on all leaves, for any sequence of handle insertions, value edits, visibility toggles, withdrawals, and $\Seal$ transactions on $H$ slots and $N = \card{\Omega}$ worlds:
\begin{enumerate}
  \item \textbf{Exactness:} In real arithmetic, the root $(n^{(1)}, b^{(1)})$ of Algorithm~\ref{alg:segtree} satisfies the unified ledger invariant~\eqref{eq:ledger-invariant}, $(n^{(1)}, b^{(1)}) = K_{\mathrm{base}} + \sum_{i=1}^{H} E_i(S)$, at every step.
  \item \textbf{Complexity:} Each single-slot update performs $N\lceil\log_2(H+1)\rceil$ integer additions and $N\lceil\log_2(H+1)\rceil$ floating-point additions, using $O(NH)$ space.
  \item \textbf{Fixed-tree history independence:} Fix the tree shape, padding, current encoded leaf values $(K_{\mathrm{base}}, E_1(S), \ldots, E_H(S))$, and IEEE-754 floating-point environment. Provided no addition or integer count overflows, the root is bitwise identical to a full rebuild on those leaves using the same tree. Inserting and then withdrawing a transient factor leaves zero residual drift relative to that rebuild, while simultaneously active factors retain ordinary tree-addition rounding.
\end{enumerate}
\end{theorem}

\begin{proof}
In real arithmetic, induction on tree height shows that every node stores the exact sum of its descendant leaves, and each updated leaf has exactly $\lceil\log_2(H+1)\rceil$ ancestors; with $2M-1 < 4(H+1)$ nodes of $2N$ scalars each, space is $O(NH)$. Because leaves $M$ and $M+i$ store $K_{\mathrm{base}}$ and $E_i(S)$ while padding leaves store $(0,0)$, the root equals $K_{\mathrm{base}}+\sum_{i=1}^H E_i(S)$. Under IEEE-754 rounding, each internal node is the deterministic machine sum of its two current children in fixed tree order. Recomputing the ancestors of each modified leaf bottom up restores this node invariant because all untouched subtrees keep identical leaves, yielding the same root bit for bit as a full rebuild on the current leaves.
\end{proof}

Table~\ref{tab:agg-comparison} compares full ledger reconstruction, direct subtraction caching, and Segment-Tree SoftREU across time, space, and floating-point behavior.

\begin{table}[htbp]
\centering
\small
\caption{Comparison of aggregation strategies for $m = \card{R\cap V}$ active visible editable records, slot capacity $H$, and $N = \card{\Omega}$ worlds.}
\label{tab:agg-comparison}
\begin{tabularx}{\linewidth}{@{}lcccX@{}}
\toprule
Strategy & Update time & Extra space & History-independent? & Floating-point behavior \\
\midrule
Full rebuild & $O(mN)$ & $O(N)$ & Yes & No subtraction drift, linear scan per edit \\
Direct cache (REU) & $O(N)$ & $O(N)$ & No & Vulnerable to binary64 cancellation drift \\
Segment-Tree SoftREU & $O(N\log H)$ & $O(NH)$ & Yes & Zero subtraction drift; exact zero recovery \\
\bottomrule
\end{tabularx}
\end{table}

\subsection{Atomic Block Deltas and Safe Folding}

The same group and tree structure handles coordinated multi-handle updates and irreversible record folding.

\paragraph{Atomic block group deltas.}
Coordinated operations such as batch channel recalibration, dependent-block withdrawal, or group publication modify a subset $\mathcal B \subseteq \mathcal H$ of handles simultaneously. Under the unified invariant~\eqref{eq:ledger-invariant}, any such transition has canonical group delta
\[
  \Delta_{\mathcal B} = \sum_{i\in\mathcal B}\bigl(E_i(S') - E_i(S)\bigr) \in (\Z\times\R)^{\Omega}.
\]
In Algorithm~\ref{alg:segtree}, for any nonempty $\mathcal B \subseteq \mathcal H$, updating the leaves of $\mathcal B$ and refreshing their shared ancestor union touches at most $O\bigl(\card{\mathcal B}\log(H/\card{\mathcal B}+1)\bigr)$ internal nodes per world; publishing the root after the entire block finishes exposes no intermediate state. Fixed replacements on disjoint handle blocks produce the same final leaves regardless of sequential order, while concurrent writes to shared ancestors require synchronization or atomic publication of a root snapshot.

\paragraph{Safe folding ($\Seal$).}
An editable visible handle $i \in R\cap V$ can be folded into $K_{\mathrm{base}}$ when the revision contract guarantees that handle $i$ is permanently immutable---neither withdrawn, edited, recalibrated, nor hidden under any allowed future operation. Under this condition, $\Seal(i)$ executes as a single atomic transaction:
\begin{enumerate}
  \item Update the base aggregate $K'_{\mathrm{base}} \leftarrow K_{\mathrm{base}} + E_i(S) = K_{\mathrm{base}} + \kappa(F_i)$ at leaf $M$. Because $K_{\mathrm{base}}\in(\N\times\R)^\Omega$ stores summed zero multiplicities and positive log sums, it accommodates multiple folded zeros at any world.
  \item Remove $i$ from $R$ (marking $i$ sealed in $\nu$ so subsequent edits targeting $i$ are rejected by $\pi_{\mathrm{rec}}$), release payload $D_i$, and reset leaf $M+i$ to $E_i(S') = (0,0)$, refreshing the ancestors of leaves $M$ and $M+i$ (at most $2\lceil\log_2(H+1)\rceil - 1$ internal nodes per world).
\end{enumerate}
By associativity of $(\N\times\R)^\Omega$, $K'_{\mathrm{base}} + \sum_{h\in\mathcal H} E_h(S') = K_{\mathrm{base}} + \sum_{h\in\mathcal H} E_h(S)$, preserving invariant~\eqref{eq:ledger-invariant} in exact arithmetic. Equivalence extends by induction across all subsequent allowed edits and reads because every remaining editable handle keeps the same effective contribution and any future request targeting the sealed handle $i$ is rejected. Releasing $D_i$ bounds editable payload storage to handles that remain mutable, while $\nu$ and $J$ retain protocol metadata as required.

In binary64 arithmetic, folding reassociates additions and changes rounding. For example, four leaf logs $(-1,0,-2^{-53},-2^{-53})$ in slots $0$--$3$ give a balanced-tree root of $(-1+0)+(-2^{-53}-2^{-53}) = -1-2^{-52}$. Folding slot~2 into the base rounds $\operatorname{fl}(-1-2^{-53})$ to $-1$ under round-to-nearest ties-to-even, and the remaining $-2^{-53}$ then rounds away at the root, yielding $-1$. To bound this reassociation change for fixed encoded logs $b_{iz}$ (expanding $K_{\mathrm{base}}$ into the prior and folded terms that formed it), let $D$ be the maximum number of additions along any term's reduction path, let $u$ be unit roundoff, and assume no overflow or underflow. By standard rounding-error analysis~\cite[Lemma~3.1]{Higham_2002},
\[
  \bigl|\widehat b_z - b_z\bigr| \le \gamma_D \sum_i \card{b_{iz}},
  \qquad \gamma_D = \frac{Du}{1-Du}, \qquad Du < 1,
\]
since each term is multiplied by at most $D$ factors $(1+\delta)$ with $\card{\delta}\le u$. If the pre- and post-fold trees have maximum path depths $D_{\mathrm{pre}}$ and $D_{\mathrm{post}}$, the triangle inequality bounds their log-sum difference by $(\gamma_{D_{\mathrm{pre}}} + \gamma_{D_{\mathrm{post}}})\sum_i \card{b_{iz}}$. Because the integer conflict counts $n_z$ are exact (absent integer overflow), combining these bounds with Proposition~\ref{prop:perturbation-stability} translates aggregation rounding directly into posterior total-variation error for the exact readout~\eqref{eq:softreu-read}:
\[
  \|\widehat p_\beta - p_\beta\|_{\mathrm{TV}}
  \le \tanh\!\bigl(\tfrac{1}{2}\gamma_D \max_{z\in\Omega}\sum\nolimits_i \card{b_{iz}}\bigr),
  \qquad
  \|\widehat p^{\mathrm{post}}_\beta - \widehat p^{\mathrm{pre}}_\beta\|_{\mathrm{TV}}
  \le \tanh\!\bigl(\tfrac{1}{2}(\gamma_{D_{\mathrm{pre}}} + \gamma_{D_{\mathrm{post}}})\max_{z\in\Omega}\sum\nolimits_i \card{b_{iz}}\bigr),
\]
uniformly over all $\beta > 0$ and all $n \in \N^\Omega$ (and identically for every query marginal $\SoftRead_{q,\beta}$). Retaining sealed leaves in a secondary balanced tree reduces the sealed terms' path depth from linear in the number of folds to logarithmic.
\FloatBarrier

\section{Commit-Gated Differentiable Learning and Gradient Alignment}\label{sec:learning}

When an evidence ledger is embedded inside a learned memory architecture such as MMLA~\cite{mmla-v4}, learned components parameterize two interfaces: (i) observation channels (such as noisy value channels $B_\theta$ and exclusion channels $C_\theta$) mapping record payloads $D_i$ to likelihood factors $F_{i,\theta}(z)$, and (ii) a candidate-addressing scorer $w_\phi(a \mid \xi) \in \Delta(\mathcal A)$ over a finite candidate edit set $\mathcal A$ that ranks which concrete edit an ambiguous revision cue $\xi$ intends.

Overwriting ledger slots in place with continuous weights $w_\phi(a\mid\xi)$ violates two properties of revisable evidence: attenuating multiple candidate slots mixes their contributions into surviving entries, so a later authoritative withdrawal of the confirmed handle leaves residual distortion on the remaining slots; and a continuous probability score is not an authenticated grant to mutate a live record. Below we separate commit-time soundness from gradient direction, then analyze when a straight-through score update agrees with the gradient of expected committed loss.

\subsection{Commit-Gated Training Interface and Soundness}

\begin{algorithm}[Commit-Gated SoftREU Training and Inference]\label{alg:commit-gated}
Given an admissible predecessor state $S = (C, K_{\mathrm{base}}, K)$, disjoint channel and addressing parameters $(\theta, \phi)$, and an annealed inverse-temperature schedule $\{\beta_t\}_{t=1}^T$ with $\beta_t \uparrow \beta_{\max}$:
\begin{enumerate}
  \item \textbf{Discrete commit / forward path (inference and authorized edits):} Execute the record transition $\pi_{\mathrm{rec}}(S, e)$ to verify delivery ID, handle generation and status in $\nu$, and authenticated grant. On acceptance at handle $i$, compute the post-edit effective contribution $E_i(S')$ from the updated record state and parameterized channels $\theta$ (taking $(0,0)$ if $i$ is hidden, withdrawn, or sealed), update leaf $M+i$ in the segment tree (Algorithm~\ref{alg:segtree}), and read $p_{q,\beta_t} = \SoftRead_{q,\beta_t}(K')$.
  \item \textbf{Isolated branch previews (ambiguous cues):} For an ambiguous cue $\xi$ over candidate edits $\mathcal A$ that passed the gate, evaluate each candidate's preview aggregate $K_a$ without modifying $S$ or consuming delivery IDs (for a single-slot candidate $a$ targeting handle $i_a$, $K_a = K - E_{i_a}(S) + E_{i_a}(S_a)$ in exact arithmetic, evaluated subtraction-free in $O(N\log H)$ additions by summing sibling nodes along the root path of leaf $M+i_a$), and form the detached score mixture
  \[
    K_{\mathrm{score}} = \sum_{a\in\mathcal A} w_\phi(a\mid\xi)\operatorname{sg}(K_a),
  \]
  where $\operatorname{sg}$ stops gradients. (Evaluating $\SoftRead_{q,\beta_t}\!\bigl(\sum_{a\in\mathcal A} w_\phi(a\mid\xi) K_a\bigr)$ yields a relaxed preview readout, whereas Step~3 evaluates the loss on a discrete branch.)
  \item \textbf{Backward path (STE + annealed SoftREU):} Let $a^*\in\mathcal A$ be the discrete branch selected for rollout (for example, the highest-scoring candidate in a sandbox or the branch bound by clarification). Attach the addressing gradient to its readout via
  \[
    \widehat{K}_{\mathrm{STE}} = K_{a^*} + K_{\mathrm{score}} - \operatorname{sg}(K_{\mathrm{score}}),
    \qquad
    \hat{p}_{q,\beta_t} = \SoftRead_{q,\beta_t}(\widehat{K}_{\mathrm{STE}}).
  \]
  For any differentiable loss $\mathcal L(\hat{p}_{q,\beta_t}, y^*)$, backpropagating via Proposition~\ref{prop:gradients} (and, when needed, the smooth encoding~\eqref{eq:soft-zero}) sends $\theta$ the derivative solely through the selected branch $K_{a^*}$ and sends $\phi$ the straight-through derivative through $K_{\mathrm{score}}$.
\end{enumerate}
\end{algorithm}

\begin{proposition}[Commit-time soundness of Algorithm~\ref{alg:commit-gated}]\label{prop:commit-soundness}
Starting from an admissible state $S$ satisfying the unified ledger invariant~\eqref{eq:ledger-invariant} with atomic enforcement of $\pi_{\mathrm{rec}}$, every live commit (and every valid $\Seal$ transaction) preserves admissibility and authorization, and no training preview or interpolation mutates the live ledger. For fixed $\theta$ and hard-encoded inference factors $\kappa(F_{i,\theta})$, the tempered read in exact arithmetic converges as $\beta\to\infty$ to the exact REU read on each consistent prefix and to the minimum-conflict distribution on each zero-mass prefix.
\end{proposition}

\begin{proof}
Because $K_{\mathrm{score}} - \operatorname{sg}(K_{\mathrm{score}})$ vanishes in the forward pass, $\widehat K_{\mathrm{STE}} = K_{a^*}$, and all branch previews execute on isolated copies without modifying $S$ or $J$. On any live transition, $\pi_{\mathrm{rec}}$ verifies delivery uniqueness, generation/seal status, and grant before updating leaf $M+i$ to the post-edit effective contribution $E_i(S')$ in Algorithm~\ref{alg:segtree} (or updating leaves $M$ and $M+i$ atomically under $\Seal(i)$), preserving equation~\eqref{eq:ledger-invariant} by induction over live commits. The $\beta\to\infty$ limit follows from Theorems~\ref{thm:beta-limit} and~\ref{thm:segtree} applied to the hard-encoded aggregate.
\end{proof}

\subsection{Expected Committed Loss and Straight-Through Curvature Bound}

Proposition~\ref{prop:commit-soundness} guarantees authorization safety, not gradient alignment. To characterize what Step~3 of Algorithm~\ref{alg:commit-gated} optimizes, fix an admissible predecessor $S$, a finite set $\mathcal A$ of gate-admitted candidate edits, and a finite continuation distribution $\mu(c)$ supplied by the task. For each branch $a\in\mathcal A$ and continuation $c$, replay $c$ on an isolated copy of the successor state $S_a$ under that branch's own $\pi_{\mathrm{rec}}$ checks (so a later request may succeed on one branch and be rejected as inactive on another), incurring a finite loss $\ell_{a,c}$ and expected branch loss $L_a = \sum_c \mu(c)\ell_{a,c}$.

Parameterize candidate scores by logits $\alpha\in\R^{\mathcal A}$ with $w = \softmax(\alpha) \in \Delta(\mathcal A)$ (writing $w_a = w_\phi(a\mid\xi)$). A randomized authorized policy commits $a\sim w$ and incurs the \emph{expected committed loss}
\begin{equation}
  R(\alpha) = \sum_{a\in\mathcal A} w_a L_a,
  \qquad
  \frac{\partial R}{\partial \alpha_j} = w_j\bigl(L_j - R(\alpha)\bigr).
  \label{eq:expected-risk}
\end{equation}
When all branches in $\mathcal A$ can be evaluated, equation~\eqref{eq:expected-risk} gives the exact score gradient directly. On sampled legal rollouts $a\sim w$, the score-function estimator $(\ell_{a,c} - u)(\mathbf{e}_a - w)$ is unbiased for $\nabla_\alpha R$ for any baseline $u$ independent of $a$~\cite{Williams_1992}.

Now fix one continuation with tempered branch logits $s_a = b_a - \beta n_a \in \R^\Omega$ and a singleton target outcome $y\in\Omega$, so $L_a = \ell(s_a) = \LSE(s_a) - (s_a)_y$. Because $(n,b)\mapsto b-\beta n$ is linear, Step~3 of Algorithm~\ref{alg:commit-gated} yields the detached logit mixture $\bar s = \sum_{a\in\mathcal A} w_a \operatorname{sg}(s_a)$ and the score-gradient surrogate
\begin{equation}
  g^{\mathrm{ST}}_j = w_j \bigl\langle \nabla\ell(s_{a^*}),\, s_j - \bar s \bigr\rangle.
  \label{eq:ste-score-grad}
\end{equation}
The next theorem gives the exact discrepancy between~\eqref{eq:expected-risk} and~\eqref{eq:ste-score-grad} and a curvature test guaranteeing that they agree in sign.

\begin{theorem}[Curvature remainder and sign-agreement criterion for commit-gated STE]\label{thm:ste-curvature}
Fix a continuation with singleton outcome $y\in\Omega$, finite branch logits $s_a = b_a - \beta n_a$, and log losses $L_a = \ell(s_a) = \LSE(s_a) - (s_a)_y$ for $a\in\mathcal A$. For any selected branch $a^*\in\mathcal A$, set $d_a = s_a - s_{a^*}$,
\[
  r_a = \ell(s_a) - \ell(s_{a^*}) - \langle\nabla\ell(s_{a^*}),\, d_a\rangle,
  \qquad
  B_a = \tfrac{1}{8}\bigl(\max_{z\in\Omega} d_a(z) - \min_{z\in\Omega} d_a(z)\bigr)^2.
\]
Then for every candidate $j\in\mathcal A$,
\begin{equation}
  \frac{\partial R}{\partial \alpha_j} - g^{\mathrm{ST}}_j
  = w_j\Bigl(r_j - \sum_{a\in\mathcal A} w_a r_a\Bigr),
  \qquad
  0 \le r_a \le B_a.
  \label{eq:ste-bias}
\end{equation}
Consequently, whenever the committed-loss advantage satisfies
\begin{equation}
  \bigl|L_j - R(\alpha)\bigr| > \max\!\Bigl((1-w_j)B_j,\; \sum_{a\neq j} w_a B_a\Bigr),
  \label{eq:sign-agreement}
\end{equation}
the straight-through component $g^{\mathrm{ST}}_j$ has the same strict sign as the exact component $\partial R/\partial\alpha_j$.
\end{theorem}

\begin{proof}
Expanding $\ell(s_a) = \ell(s_{a^*}) + \langle\nabla\ell(s_{a^*}), s_a - s_{a^*}\rangle + r_a$ inside $\partial R/\partial\alpha_j = w_j\bigl(\ell(s_j) - \sum_a w_a\ell(s_a)\bigr)$ cancels $\ell(s_{a^*})$; the linear term equals $w_j\langle\nabla\ell(s_{a^*}), s_j - \bar s\rangle = g^{\mathrm{ST}}_j$, leaving the first identity in~\eqref{eq:ste-bias}. Along the segment $s_{a^*} + t d_a$ ($t\in[0,1]$), put $\psi(t) = \ell(s_{a^*} + t d_a)$. Direct differentiation gives $\psi'(t) = \langle\softmax(s_{a^*}+t d_a) - \mathbf{e}_y, d_a\rangle$ and $\psi''(t) = \operatorname{Var}_{\softmax(s_{a^*}+t d_a)}(d_a) \ge 0$. By Popoviciu's inequality, any distribution on $\Omega$ satisfies $\operatorname{Var}(d_a) \le \frac{1}{4}\bigl(\max_z d_a(z) - \min_z d_a(z)\bigr)^2 = 2B_a$. Taylor's theorem with integral remainder therefore gives
\[
  0 \le r_a = \int_0^1 (1-t)\psi''(t)\,dt \le 2B_a\int_0^1 (1-t)\,dt = B_a.
\]
Writing $r_j - \sum_a w_a r_a = (1-w_j)r_j - \sum_{a\neq j} w_a r_a$ and applying $0\le r_a\le B_a$ gives $-\sum_{a\neq j} w_a B_a \le r_j - \sum_a w_a r_a \le (1-w_j)B_j$. Under~\eqref{eq:sign-agreement}, $\card{r_j - \sum_a w_a r_a} < \card{L_j - R(\alpha)}$, so $g^{\mathrm{ST}}_j/w_j = (L_j - R(\alpha)) - (r_j - \sum_a w_a r_a)$ shares the strict sign of $L_j - R(\alpha)$; since $w_j > 0$, $g^{\mathrm{ST}}_j$ shares the sign of $\partial R/\partial\alpha_j$.
\end{proof}

Under a finite continuation law $\mu(c)$ with singleton outcomes, linearity averages $L_a$, $r_a$, and $B_a$ with respect to $\mu(c)$ and equations~\eqref{eq:ste-bias}--\eqref{eq:sign-agreement} hold unchanged. For a multi-world query outcome $q^{-1}(y)$ with $\card{q^{-1}(y)}\ge 2$, the negative log probability $-\log\SoftRead_{q,\beta}(y) = \LSE(s) - \LSE_{z\in q^{-1}(y)}(s)$ is a difference of two log-sum-exp functions whose directional curvature equals the difference of an unconditional and a conditional variance, so nonnegativity of $r_a$ uses the singleton assumption.

\paragraph{Two-action bias identity.}
For two legal actions $\mathcal A = \{0,1\}$ with $w_1 = \sigma(\phi)$, $w_1' = w_1(1-w_1)$, $s(t) = (1-t)s_0 + t s_1$, and $f(t) = \ell(s(t))$, the exact scalar derivative is $R'(\phi) = w_1'[f(1) - f(0)]$, whereas the straight-through surrogate is $g^{\mathrm{ST}} = w_1' f'(a^*)$ for $a^*\in\{0,1\}$. Integration by parts gives the exact estimator bias $g^{\mathrm{ST}} - R'(\phi)$:
\[
  g^{\mathrm{ST}} - R'(\phi) = \begin{cases}
    +w_1' \displaystyle\int_0^1 t\,f''(t)\,dt \ge 0, & a^* = 1,\\[2.2ex]
    -w_1' \displaystyle\int_0^1 (1-t)\,f''(t)\,dt \le 0, & a^* = 0.
  \end{cases}
\]
Because $f''(t) \ge 0$, evaluating the local slope at an endpoint systematically overshoots the secant slope in opposite directions at $a^*=1$ and $a^*=0$, and can reverse the update sign whenever the secant gap $f(1)-f(0)$ is smaller than the integrated curvature.

\subsection{Exact Sign Reversal on Authorized Withdrawals and Continuation Flip}

The following construction shows that sign reversal occurs on an admissible ledger with strictly positive likelihoods, exact arithmetic, and zero dependence on $\beta$.

\begin{proposition}[Positive-support sign reversal and continuation preference flip]\label{prop:ste-reversal}
Let $\Omega = \{00, 10, 01\}$ with uniform prior $\rho = (1/3, 1/3, 1/3)$ and two visible live records $r_0, r_1$ carrying positive binary-channel factors
\[
  F_0 = (1/2,\, 1/2,\, 1/8), \qquad F_1 = (1/2,\, 1/32,\, 1/2).
\]
Consider two authorized withdrawals: $a=0$ withdraws $r_1$, and $a=1$ withdraws $r_0$, evaluated on singleton outcome $y = 00$ with $w_1 = \sigma(\phi)$. Then:
\begin{enumerate}
  \item \textbf{Exact sign reversal:} Because $n_z = 0$ on every world and branch, $s_a = b_a$ is independent of $\beta > 0$, and $R'(\phi) = w_1(1-w_1)\log(11/12) < 0$ for all $\phi\in\R$. Yet the deterministic argmax STE ($a^* = 1$) satisfies $g^{\mathrm{ST}}_{\mathrm{argmax}} > 0$ for every $w_1 \in (1/2, 1)$, and the sampled-branch expected STE ($a^*\sim w$) satisfies $\mathbb{E}[g^{\mathrm{ST}}] > 0$ for every $w_1 \in (11/17, 1)$. In particular, at $w_1 = 3/4$ ($\phi = \log 3$),
  \[
    R'(\log 3) = \tfrac{3}{16}\log(11/12) < 0, \qquad
    g^{\mathrm{ST}}_{\mathrm{argmax}} = \tfrac{7}{44}\log 2 > 0, \qquad
    \mathbb{E}_{a^*\sim w}\bigl[g^{\mathrm{ST}}\bigr] = \tfrac{49}{1056}\log 2 > 0.
  \]
  \item \textbf{Continuation preference flip:} Suppose that with probability $\eta \in [0,1]$, a fresh authorized request withdraws $r_1$ at the common new version before the query is read (and with probability $1-\eta$, the query is read immediately). Then the expected loss difference $L_1(\eta) - L_0(\eta) = (1-\eta)\log(11/12) + \eta\log(4/3)$ changes sign at $\eta^* = \log(12/11)/\log(16/11) \approx 0.2322$; in particular, branch $a=0$ becomes strictly preferred at $\eta = 1/4$.
\end{enumerate}
\end{proposition}

\begin{proof}
All entries of $F_0, F_1$ and their complements lie in $[0,1]$, so both are valid binary observation channels. Table~\ref{tab:ste-trace} lists the exact posterior probabilities computed from the unified invariant~\eqref{eq:ledger-invariant}: $p_0 = (4/9, 4/9, 1/9)$ after withdrawing $r_1$, and $p_1 = (16/33, 1/33, 16/33)$ after withdrawing $r_0$. Thus $L_1 - L_0 = -\log(16/33) + \log(4/9) = \log(11/12) < 0$, giving $R'(\phi) = w_1(1-w_1)\log(11/12) < 0$.

The logit difference is $d = s_1 - s_0 = (0, -4, 2)\log 2$. Since $d(00) = 0$, the directional derivative $\langle p_k - \mathbf{e}_{00}, d\rangle$ equals $\mathbb{E}_{p_k}[d]$:
\[
  f'(0) = \tfrac{4(-4) + 1(2)}{9}\log 2 = -\tfrac{14}{9}\log 2, \qquad
  f'(1) = \tfrac{1(-4) + 16(2)}{33}\log 2 = \tfrac{28}{33}\log 2 > 0.
\]
For $w_1 > 1/2$, $a^* = \arg\max_a w_a = 1$, so $g^{\mathrm{ST}}_{\mathrm{argmax}} = w_1(1-w_1)f'(1) > 0$. Under sampled rollouts $a^*\sim w$, $\mathbb{E}[g^{\mathrm{ST}}] = w_1(1-w_1)\bigl((1-w_1)f'(0) + w_1 f'(1)\bigr) > 0$ holds whenever $\frac{28}{33} w_1 > \frac{14}{9}(1-w_1)$, i.e., $w_1 > 11/17$. Evaluating at $w_1 = 3/4$ ($w_1' = 3/16$) gives $g^{\mathrm{ST}}_{\mathrm{argmax}} = \frac{3}{16}\cdot\frac{28}{33}\log 2 = \frac{7}{44}\log 2$ and $\mathbb{E}[g^{\mathrm{ST}}] = \frac{3}{16}\bigl(-\frac{7}{18}+\frac{7}{11}\bigr)\log 2 = \frac{49}{1056}\log 2$.

For the continuation, a fresh authorized withdrawal of $r_1$ finds $r_1$ already retired on branch $a=0$ (preserving $p(00) = 4/9$) and succeeds on branch $a=1$ (retiring the remaining record and restoring the uniform prior $p(00) = 1/3$). The post-continuation loss difference is $-\log(1/3) + \log(4/9) = \log(4/3) > 0$, yielding $L_1(\eta) - L_0(\eta) = (1-\eta)\log(11/12) + \eta\log(4/3)$. Setting this to zero gives $\eta^* = \log(12/11)/\log(16/11)$. At $\eta = 1/4$, $(11/12)^3(4/3) = 1331/1296 > 1$, so $L_1(1/4) - L_0(1/4) = \frac{1}{4}\log(1331/1296) > 0$.
\end{proof}

\begin{table}[htbp]
\centering
\small
\caption{Exact query probabilities for the positive-support witness of Proposition~\ref{prop:ste-reversal}. Both previews operate on isolated copies of the intact initial state before branch $a=1$ commits and a fresh request withdraws $r_1$.}
\label{tab:ste-trace}
\begin{tabular}{@{}lll@{}}
\toprule
Operation & Live editable records $R$ & Exact $p(00)$ \\
\midrule
Initial state & $\{r_0, r_1\}$ & $16/21$ \\
Isolated preview $a=0$ (withdraw $r_1$) & $\{r_0\}$ & $4/9$ \\
Isolated preview $a=1$ (withdraw $r_0$) & $\{r_1\}$ & $16/33$ \\
Live commit $a=1$ & $\{r_1\}$ & $16/33$ \\
Fresh withdrawal of $r_1$ (on branch $a=1$ / branch $a=0$) & $\varnothing$ / $\{r_0\}$ & $1/3$ / $4/9$ \\
\bottomrule
\end{tabular}
\end{table}

Proposition~\ref{prop:ste-reversal} clarifies the operational boundary between Algorithm~\ref{alg:commit-gated}'s STE surrogate and exact expected-loss updates: STE is fast and sign-consistent when branch-logit oscillations satisfy~\eqref{eq:sign-agreement}, whereas regimes with large branch curvature or continuation-dependent liveness flips require exact branch enumeration or unbiased score-function gradients~\eqref{eq:expected-risk}.
\FloatBarrier

\section{Verification and Numerical Experiments}\label{sec:experiments}

We verify every analytical bound, gradient identity, protocol invariant, sign-reversal calculation, and complexity claim using three self-contained ancillary scripts: \texttt{anc/checks/check\_softreu.py} (numerical checks and channel fit), \texttt{anc/checks/check\_protocol\_and\_ste.py} (exact-rational protocol and STE witness, Python standard library only), and \texttt{anc/checks/benchmark\_scaling.py} (single-slot latency scaling, Python standard library only).

\subsection{Exact \texorpdfstring{$\beta$}{beta}-Convergence and Minimum-Conflict Fallback}

On a four-world ledger with zero counts $n = (0, 1, 0, 2)$ and positive log masses $b = \log(1, 3, 2, 4)$, the exact REU posterior is $p_\infty = (1/3, 0, 2/3, 0)$ with $\Delta n_{\min} = 1$ and $\sum_{z\notin U_{\min}} e^{b_z} / \sum_{u\in U_{\min}} e^{b_u} = 7/3$. Table~\ref{tab:beta-conv} and Figure~\ref{fig:experiments}(a) confirm that the empirical $L_1$ error $\|p_\beta - p_\infty\|_1$ decays exponentially in $\beta$ and remains strictly below the Theorem~\ref{thm:beta-limit} upper bound $(14/3)\,e^{-\beta}$ across every tested temperature, reaching $2.25\times 10^{-7}$ at $\beta=16$.

On a conflicting two-world zero-mass ledger with $n = (1, 2)$ and $b = \log(1/2, 1/2)$, exact REU returns $\bot$. SoftREU remains well-defined and concentrates smoothly on the minimum-conflict world $z=0$ ($n_0 = 1 < n_1 = 2$), giving $p_{10}(0) = (1+e^{-10})^{-1} \approx 0.999955$ and $p_\infty = (1, 0)$.

\begin{table}[htbp]
\centering
\small
\caption{Empirical $L_1$ convergence of SoftREU to exact REU versus the analytical upper bound of Theorem~\ref{thm:beta-limit} on $n=(0,1,0,2)$, $b=\log(1,3,2,4)$.}
\label{tab:beta-conv}
\setlength{\tabcolsep}{4.2pt}\begin{tabular}{@{}lcccccccc@{}}
\toprule
Inverse temperature $\beta$ & $0.5$ & $1.0$ & $2.0$ & $4.0$ & $6.0$ & $8.0$ & $12.0$ & $16.0$ \\
\midrule
Empirical $\|p_\beta - p_\infty\|_1$ & $1.05\times 10^{0}$ & $7.08\times 10^{-1}$ & $2.75\times 10^{-1}$ & $3.68\times 10^{-2}$ & $4.96\times 10^{-3}$ & $6.71\times 10^{-4}$ & $1.23\times 10^{-5}$ & $2.25\times 10^{-7}$ \\
Theorem~\ref{thm:beta-limit} bound & $2.83\times 10^{0}$ & $1.72\times 10^{0}$ & $6.32\times 10^{-1}$ & $8.55\times 10^{-2}$ & $1.16\times 10^{-2}$ & $1.57\times 10^{-3}$ & $2.87\times 10^{-5}$ & $5.25\times 10^{-7}$ \\
\bottomrule
\end{tabular}
\end{table}

\subsection{Gradient Check, Monotonicity Boundary, and Binary64 Stability}

\paragraph{Analytic Jacobians and boundary checks.}
We evaluate equations~\eqref{eq:grad-b}--\eqref{eq:grad-beta} on $n=(0,1,0,2)$, $b=(-0.5, 0.7, 0.3, 1.1)$, binary query $q=(0,0,1,1)$, and $\beta=2.5$ against central finite differences (step size $h = 10^{-6}$). All three gradients ($\partial/\partial b$, $\partial/\partial n$, and $\partial/\partial\beta$) agree to within $10^{-9}$. The script also checks the Remark~\ref{rem:soft-zero-mono} boundary at $\beta=0.1, \tau=0.08$ (confirming $s_\beta(10^{-6}) < s_\beta(0)$ when $\epsilon = 0.5 < e^{-0.1}$ and $s_\beta(10^{-6}) > s_\beta(0)$ when $\epsilon = e^{-0.1}$) and the binary64 $\Seal$ reassociation example of Section~\ref{sec:segtree} (returning $-1-2^{-52}$ before folding and $-1$ after folding).

\paragraph{Adversarial binary64 withdrawal.}
On two equally likely worlds with a surviving factor $\log F_2=(-1,0)$ (true posterior $p^*(0)=1/(1+e)\approx 0.268941$), inserting and withdrawing a single transient factor of log magnitude $2^{53}$ shifts the direct subtraction cache to $p(0)=0.500000$. In the 200-cycle trace of Figure~\ref{fig:experiments}(b), transient magnitudes are drawn from $\{2^{45},\ldots,2^{53}\}$ across random slots. Direct log-cache subtraction loses the surviving term at cycle~5 and retains a persistent $0.231059$ error thereafter, whereas Segment-Tree SoftREU matches the reference read (error $<10^{-15}$) at all 200 steps.

\subsection{Synthetic Binary-Channel Fit on Fixed Zero-Masked Ledgers}

\begin{figure}[htbp]
\centering
\includegraphics[width=\linewidth]{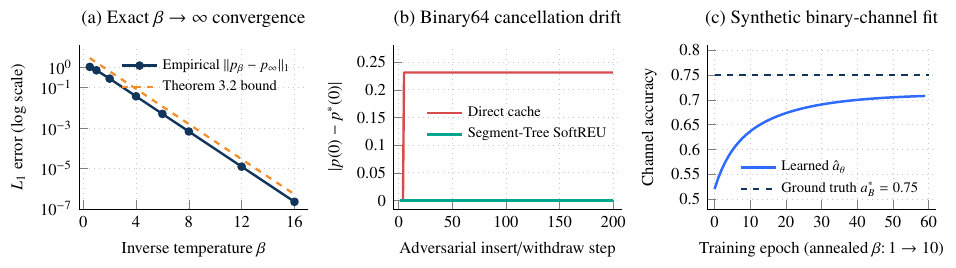}
\caption{Empirical verification of SoftREU (\texttt{anc/checks/check\_softreu.py}). (a) Exponential decay of $\|p_\beta - p_\infty\|_1$ tracked by the Theorem~\ref{thm:beta-limit} upper bound. (b) Elimination of binary64 subtraction drift across 200 adversarial insert-and-withdraw cycles by Segment-Tree SoftREU (direct cache loses the surviving term at step~5). (c) Synthetic binary-channel accuracy fit on fixed zero-masked ledgers under temperature annealing ($\beta: 1\to 10$).}
\label{fig:experiments}
\end{figure}

To test gradient flow through tempered reads on fixed zero-masked ledgers, we fit a single parameter of a symmetric binary value channel $B_\theta(v\mid x)$ with accuracy $a_\theta = \sigma(\theta)$, initialized at $a_0 = 0.520$ against a generating value $a_B^* = 0.750$. Each of 250 synthetic training episodes contains two noisy observations on fixed slots together with a $35\%$ probability of a fixed structural zero blocker on the opposing world $1-x$. Minimizing negative log-likelihood through SoftREU while linearly annealing $\beta$ from $1$ to $10$ over 60 epochs drives the learned accuracy smoothly toward the generating value ($\hat a_\theta = 0.7082$, final training NLL $0.3482$; Figure~\ref{fig:experiments}(c)).

\FloatBarrier

\subsection{Exact-Arithmetic Protocol, Unified Invariant, and STE Witness}

The standard-library checker \texttt{anc/checks/check\_protocol\_and\_ste.py} executes all finite checks in exact rational arithmetic (Python \texttt{fractions.Fraction}), cross-validating the unified aggregate read~\eqref{eq:ledger-invariant} against direct visible-factor products at every step:
\begin{itemize}
  \item \textbf{Unified invariant and $\Seal$:} On two equally likely worlds with a hidden handle initialized to $(1,1)$, editing the hidden factor to $(1/2,1)$ leaves the visible read at $(1/2,1/2)$; publishing the handle transitions the read to $(1/3,2/3)$; applying $\Seal$ moves the contribution into $K_{\mathrm{base}}$, preserves $(1/3,2/3)$, empties $R$, and causes a subsequent withdrawal request on that handle to return \texttt{sealed}.
  \item \textbf{Record gate, perturbation stability, and zero-mass recovery:} The script verifies preview isolation, denied grants, stale expected versions, duplicate delivery-ID replay (including reusing an accepted delivery ID against a different handle), recovery of a supported posterior $(1,0)$ from a conflicting two-world \texttt{zeroMass} ledger via a single named withdrawal, and Proposition~\ref{prop:perturbation-stability} (monotone conflict derivative $-\operatorname{Var}_{p_\beta}(n_Z)$ to $<10^{-9}$, exact tightness of $\tanh(\epsilon/2)$ on two worlds, and 200 randomized log-sum perturbations across $\beta\in\{0.5, 5, 50\}$).
  \item \textbf{Theorem~\ref{thm:ste-curvature} and Proposition~\ref{prop:ste-reversal}:} The script reproduces Table~\ref{tab:ste-trace}, confirms $R'(\log 3) \approx -0.016315$ against central finite differences ($h = 10^{-6}$, relative error $<10^{-9}$) alongside $g^{\mathrm{ST}}_{\mathrm{argmax}} \approx +0.110273$ and $\mathbb{E}[g^{\mathrm{ST}}] \approx +0.032163$, verifies the curvature remainder $r_0 \approx 0.6751 \le B_0 \approx 2.1620$ with $|L_1 - R| \approx 0.0218 < w_0 B_0 \approx 0.5405$ (consistent with sign reversal), and verifies $(11/12)^3(4/3) = 1331/1296 > 1$ at $\eta = 1/4$. Figure~\ref{fig:ste-scaling}(a) plots the three derivatives over $w_1\in(0,1)$, highlighting the region $w_1 > 11/17$ where both STE variants point uphill.
\end{itemize}

\subsection{Single-Slot Update Scaling Across Handle Capacity}

\begin{figure}[htbp]
\centering
\includegraphics[width=\linewidth]{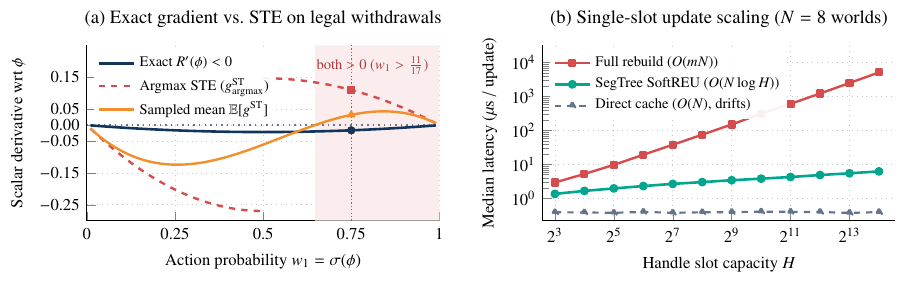}
\caption{Gradient alignment and update scaling. (a) Exact expected-loss derivative $R'(\phi)$ versus deterministic argmax STE and sampled-branch mean STE on the positive-support witness of Proposition~\ref{prop:ste-reversal}; markers indicate the witness $w_1=3/4$, and the shaded region $w_1>11/17$ marks simultaneous sign reversal of both surrogates. (b) Median single-slot update latency (pure Python, $N=8$ worlds, \texttt{anc/checks/scaling\_results.json}) across handle capacities $H=2^3$ to $2^{14}$.}
\label{fig:ste-scaling}
\end{figure}

To verify the $O(N\log H)$ complexity of Theorem~\ref{thm:segtree} empirically, \texttt{anc/checks/benchmark\_scaling.py} times random single-slot updates in pure Python~3.12 at $N = 8$ worlds across handle capacities $H \in \{2^3, 2^4, \ldots, 2^{14}\}$ (median of five trials, raw measurements saved in \texttt{anc/checks/scaling\_results.json}). As shown in Figure~\ref{fig:ste-scaling}(b), full rebuild latency grows linearly from $2.95\,\mu\text{s}$ at $H = 8$ to $5{,}140.96\,\mu\text{s}$ at $H = 16{,}384$, whereas Segment-Tree SoftREU grows logarithmically from $1.36\,\mu\text{s}$ to $6.25\,\mu\text{s}$ (an $823\times$ speedup at $H = 16{,}384$, roughly $0.44\,\mu\text{s}$ per added tree level) while preserving fixed-tree history independence. Direct subtraction caching remains near $0.39\,\mu\text{s}$ across $H$ at the cost of binary64 cancellation drift (Figure~\ref{fig:experiments}(b)).

\FloatBarrier

\section{Related Work, Scope, and Conclusion}\label{sec:discussion}

\paragraph{Softmax relaxations and straight-through estimators.}
Shifted softmax and log-sum-exp provide numerically stable normalization on positive supports~\cite{Blanchard_lse}, while Gumbel-Softmax~\cite{Jang_2017} and straight-through estimators~\cite{Bengio_2013} propagate gradients through discrete categorical choices. Prior analyses of straight-through bias and gradient alignment focus on binary or quantized neural activations~\cite{Yin_2019,Shekhovtsov_2021} and bias-constrained categorical estimators~\cite{Hooper_2026}, alongside unbiased score-function gradients~\cite{Williams_1992}. SoftREU addresses a distinct algebraic and operational setting: exact reversibility across structural zeros in the Grothendieck group $(\Z\times\R)^\Omega$, paired with branch losses generated by authorized named ledger transitions and branch-dependent continuation liveness checks. Theorem~\ref{thm:ste-curvature} and Proposition~\ref{prop:ste-reversal} show that straight-through sign reversal arises directly from log-sum-exp curvature across legal evidence branches even before structural zeros or floating-point errors appear.

\paragraph{Provenance, factored representations, and MMLA.}
Maintaining explicit provenance for retraction goes back to assumption-based truth maintenance~\cite{deKleer_1986}, while differentiable probabilistic evaluation over factored distributions is formalized by arithmetic circuits~\cite{Darwiche_2003}. Algorithm~\ref{alg:segtree} operates over an explicit finite world table of size $N=\card{\Omega}$; scaling SoftREU to high-dimensional target spaces requires pairing the subtraction-free zero-count group with factored representations such as arithmetic circuits~\cite{Darwiche_2003}. Within MMLA~\cite{mmla-v4}, which defines typed memory carriers and learning lifetimes independently of the sequence mixer, SoftREU supplies a differentiable, floating-point-stable readout and aggregation primitive compatible with a strictly audited record layer.

\paragraph{Scope and conclusion.}
SoftREU unifies three mechanisms under one ledger invariant: the tempered variational readout $p_\beta = \softmax(b-\beta n)$ with its smooth-encoding monotonicity characterization $\epsilon \ge e^{-\beta}$, a subtraction-free segment tree supporting $O(N\log H)$ single-slot updates, atomic block deltas, and bounded-error $\Seal$ folding, and commit-gated learning accompanied by exact curvature conditions and a constructive sign-reversal witness separating straight-through surrogates from expected committed loss. The gradient analysis assumes a finite enumerated candidate set, a supplied observation model, and reliable gate inputs; extending empirical comparisons across learned multi-step addressing policies, score-function baselines, and factored circuit backbones remains open for future work.

\bibliographystyle{plainurl}
\bibliography{softreu}

\end{document}